\documentclass[11pt]{article}

\usepackage[final]{acl}

\usepackage{times}
\usepackage{latexsym}

\usepackage[T1]{fontenc}

\usepackage[utf8]{inputenc}

\usepackage{microtype}

\usepackage{inconsolata}

\usepackage{graphicx}

\usepackage{booktabs}
\usepackage{amssymb}

\title{ViT-Explainer: An Interactive Walkthrough \\of the Vision Transformer Pipeline}

\author{Juan Manuel Hernandez\textsuperscript{1},  Mariana Fern\'andez-Espinosa\textsuperscript{2}, \\ {\bf Denis Parra}\textsuperscript{1}, {\bf Diego G\'omez-Zar\'a}\textsuperscript{2} \\ 
        \textsuperscript{1}Pontificia Universidad Cat\'olica de Chile, \textsuperscript{2}University of Notre Dame
           }

\begin{document}
\maketitle

\begin{abstract}
Transformer-based architectures have become the shared backbone of natural language processing and computer vision. However, understanding how these models operate remains challenging, particularly in vision settings, where images are processed as sequences of patch tokens. Existing interpretability tools often focus on isolated components or expert-oriented analysis, leaving a gap in guided, end-to-end understanding of the full inference pipeline. To bridge this gap, we present \textbf{ViT-Explainer}, a web-based interactive system that provides an integrated visualization of Vision Transformer inference, from patch tokenization to final classification. The system combines animated walkthroughs, patch-level attention overlays, and a vision-adapted Logit Lens within both guided and free exploration modes. A user study with six participants suggests that ViT-Explainer is easy to learn and use, helping users interpret and understand Vision Transformer behavior.
\end{abstract}

\section{Introduction}
\label{sec:intro}
Transformer-based models dominate both natural language processing and computer vision \citep{vaswani2017attention,dosovitskiy2020image}, yet their internal decision-making processes remain difficult to interpret. As these models are increasingly adopted in research, industry, and educational contexts, the lack of accessible interpretability tools limits broader understanding and effective use of these systems \citep{Zhang_2021,e23010018,teng2022survey}. Prior work shows that interpretability influences user trust, transparency, and adoption in real-world applications \citep{ALI2023101805}. 

While substantial research has examined attention heads, residual streams, feed-forward layers, and layer-wise prediction dynamics \citep{clark-etal-2019-bert,elhage2021mathematical,geva-etal-2021-transformer,belrose2023eliciting}, it remains difficult to understand how these components operate collectively across the full inference pipeline, especially when tracing how intermediate representations evolve and culminate in a final prediction.

Interpretability is especially difficult in Vision Transformers (ViTs) \citep{dosovitskiy2020image}, which transform images into patch tokens and process them through stacked Transformer encoders. Existing tools either focus on isolated components, provide expert-oriented analytical dashboards, or target NLP-based Transformers \citep{vig2019multiscale,wang2021dodrio,li2023does,ma2023visualizing,zhou2023vit}. However, none of these tools offers a guided, end-to-end walkthrough connecting image tokenization, attention dynamics, and final classification within a unified interactive environment. Moreover, visualizing ViTs poses challenges absent in text settings, since attention must be mapped onto 2D image regions rather than sequential tokens, and intermediate Logit Lens predictions are decoded into class labels rather than readable words.

To address this gap, we introduce \textbf{ViT-Explainer}, a web-based interactive system that visualizes the complete Vision Transformer inference pipeline. The system enables users to upload images, inspect patch tokenization and embedding formation, explore spatial attention overlays across heads and layers, and trace how class predictions evolve through a vision-adapted Logit Lens \citep{belrose2023eliciting}. The interface combines a structured guided walkthrough with a free exploration mode. ViT-Explainer is designed to support students learning Vision Transformers, educators teaching model internals, and researchers seeking intuitive inspection tools.

We evaluate ViT-Explainer through a preliminary user study with six participants, which indicates that the system is usable and associated with low perceived cognitive workload. Participants highlighted the usefulness of the patch-level attention visualizations and the explicit breakdown of Multi-Head Self-Attention. Spatial overlays helped them understand how individual patches distribute attention across the image, while the step-by-step decomposition of Queries, Keys, and Values (Q, K, V) projections clarified how parallel attention heads operate within each Transformer block.

In summary, our contributions are: (1) an interactive, end-to-end visualization of Vision Transformer inference that integrates patch-level attention overlays and a vision-adapted logit lens; and (2) a user evaluation demonstrating high usability and low cognitive workload.

ViT-Explainer aligns with the ACL 2026 focus on Explainability of NLP Models by extending interactive interpretability techniques to Vision Transformers and adapting Logit Lens \citep{belrose2023eliciting} analysis to the image classification setting. Beyond model inspection, the system also serves as an educational tool, supporting structured learning of Transformer internals through guided walkthroughs and interactive exploration. A live demo is available at \url{https://vit-explainer.vercel.app/}, and a video walkthrough can be found at \url{https://youtu.be/vy4beZQo89w}.

\section{Related Work}
\label{sec:related}

\subsection{Interactive Visualization for Deep Learning.}
Interactive visualization has long been used to make neural networks more accessible, from early explorations of multilayer perceptrons to systems such as CNN~Explainer \citep{wang2020cnn}, which visualizes convolutional architectures through integrated, multi-level views. More recently, tools such as Transformer Explainer \citep{cho2025transformer} and TransforLearn \citep{gao2023transforlearn} have provided interactive walkthroughs of language-based Transformer models. Building on this tradition, we extend interactive visualization to Vision Transformers. Unlike prior systems that focus on language-based Transformers or convolutional architectures, ViT-Explainer provides an integrated, end-to-end walkthrough of the full Vision Transformer inference process, explicitly connecting spatial patch representations with Transformer internals and final predictions.

\subsection{Transformer Interpretability.}
Attention visualization has been central to Transformer interpretability since \citet{vaswani2017attention}. Tools such as BertViz \citep{vig2019multiscale} and Dodrio \citep{wang2021dodrio} enable interactive exploration of attention heads, while AttentionViz \citep{yeh2023attentionviz} provides global views of query--key interactions. Complementing attention-based analysis, the logit lens technique reveals how predictions evolve across layers \citep{belrose2023eliciting}. While these systems support fine-grained inspection of individual components, they primarily facilitate analytical probing rather than a structured understanding of how computations unfold across the full model. ViT-Explainer extends these component-level interpretability techniques to the vision setting within an integrated, end-to-end walkthrough.

\subsection{Vision Transformer Analysis.}
Several systems have explored expert-oriented analysis of Vision Transformers. \citet{li2023does} identify important attention heads through pruning metrics, \citet{ma2023visualizing} examine patch interactions across layers, and EL-VIT \citep{zhou2023vit} provides a multi-layer analytics framework. Related work has also extended interpretability to multimodal and attribution settings, including cross-modal attention visualization \citep{aflalo2022vlinterpret} and customizable feature attributions \citep{boyle2025cafga}. While these systems provide detailed analytical capabilities for expert inspection, they prioritize fine-grained analysis over guided conceptual understanding of the full ViT inference process. In contrast, ViT-Explainer provides a complete, step-by-step walkthrough of the ViT pipeline designed to help users learn and analyze ViTs, integrating spatial attention overlays and layer-wise prediction tracing within a single interface.

\section{System Description}
\label{sec:system}

\begin{figure*}[t]
    \centering
    \includegraphics[width=\textwidth]{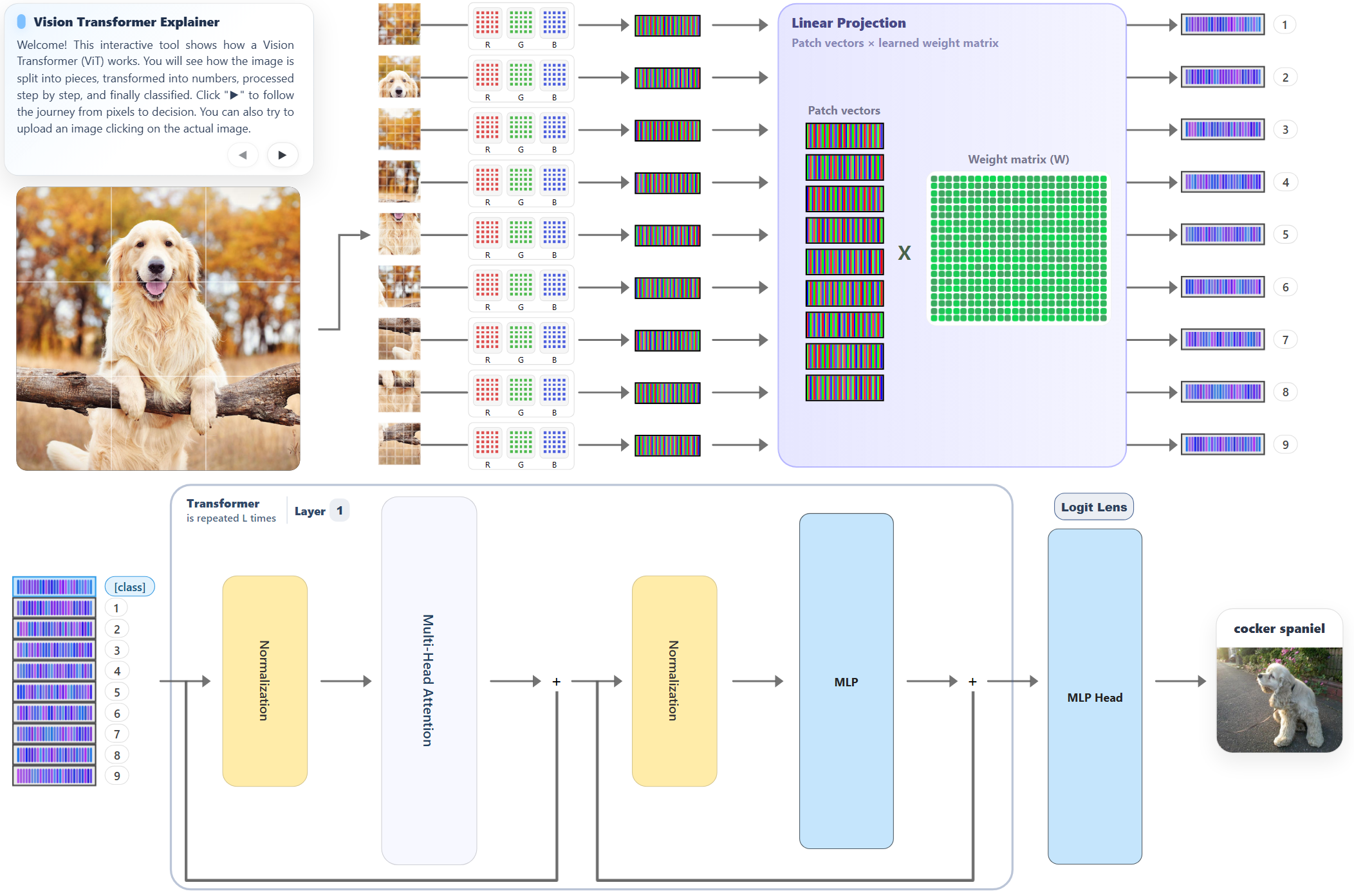}
    \caption{\textbf{ViT-Explainer} interface.
    \textbf{Top:} The input image is divided into non-overlapping patches, decomposed into RGB channels, flattened into vectors, and projected into an embedding space via a learned linear transformation. 
    \textbf{Bottom:} A Transformer encoder block is visualized step by step, including layer normalization, Multi-Head Self-Attention, residual connections, and the MLP sublayer. 
    The classification head produces logits from the class token, and users can navigate the processing stages through guided controls (top-left).}
    \label{fig:system}
\end{figure*}

ViT-Explainer is a web-based interactive visualization system designed to explain the Vision Transformer inference pipeline. It integrates model inference, intermediate representation extraction, and interactive visualizations within a unified browser interface. In this section, we describe the system architecture and the visualization pipeline.

\subsection{System Architecture}
\label{sec:architecture}

ViT-Explainer follows a client–server architecture. The \textbf{frontend} (Figure~\ref{fig:system}) is implemented in Svelte\footnote{\url{https://svelte.dev/}} and JavaScript, and deployed on Vercel\footnote{\url{https://vercel.com/}}. It renders the interactive visualization pipeline entirely in the browser using SVG-based components and reactive state management. The interface supports both a guided, step-by-step walkthrough and a free exploration mode, enabling users to interact with internal representations at their own pace.

The \textbf{backend} is implemented in Python using PyTorch and the \texttt{timm} library\footnote{\url{https://huggingface.co/docs/timm/en/index}}, and is deployed on Hugging Face Spaces. We use a pretrained FlexiViT-Large model \texttt{flexivit\_large.300ep\_in1k}, adapted to a $3\times3$ patch grid (96×96 input resolution with 32×32 patches). The model comprises 24 Transformer encoder layers with Multi-Head Self-Attention (16 heads per layer) and a hidden dimension of 1024.

We use a $3\times3$ patch configuration as a pedagogical choice for interpretability. With 9 patches plus the CLS token, the resulting $10\times10$ attention matrices can be displayed in full without scrolling or aggregation, allowing users to inspect every individual attention weight. Standard configurations (e.g., $14\times14$ patches) would produce larger matrices that cannot be easily visualized at element level in a browser interface.

Upon image upload, the backend performs a forward pass and extracts intermediate representations, including attention weights for all heads and layers, final logits, and softmax probabilities. To support the Logit Lens visualization, we apply the classification head to the class token at each layer, producing a trajectory of prediction distributions across network depth. These intermediate tensors are returned to the frontend for interactive rendering.

The frontend presents the ViT pipeline as a sequence of coordinated, animated views. In guided mode, users navigate through a structured walkthrough with forward and backward controls, where each step highlights a specific processing stage and provides contextual explanations. In free exploration mode, users can directly interact with patches, attention heads, layers, and prediction trajectories without the step-based constraints. The visual design follows Shneiderman's (\citeyear{shneiderman2003eyes}) information-seeking mantra: overview first, zoom and filter, and then details on demand.

\subsection{Visualization Pipeline}
\label{sec:pipeline}

ViT-Explainer decomposes the Vision Transformer into coordinated interactive stages that progressively reveal the internal processing pipeline.

\subsubsection{Image Patching and Tokenization.}
The input image is divided into non-overlapping patches with an overlaid grid. Patch extraction and RGB decomposition are animated to show how pixels are flattened into vectors. For illustrative purposes, the guided mode begins with a simplified $3 \times 3$ grid before transitioning to the model’s actual resolution.

\subsubsection{Linear Projection and Positional Encoding.}
Flattened patch vectors are projected into an embedding space via a learned weight matrix, visualized as matrix multiplication. Positional embeddings are added through animated element-wise addition, and a CLS token is prepended as a dedicated aggregation vector.

\subsubsection{Transformer Block Internals.}
\begin{figure}[t]
    \centering
    \includegraphics[width=\columnwidth]{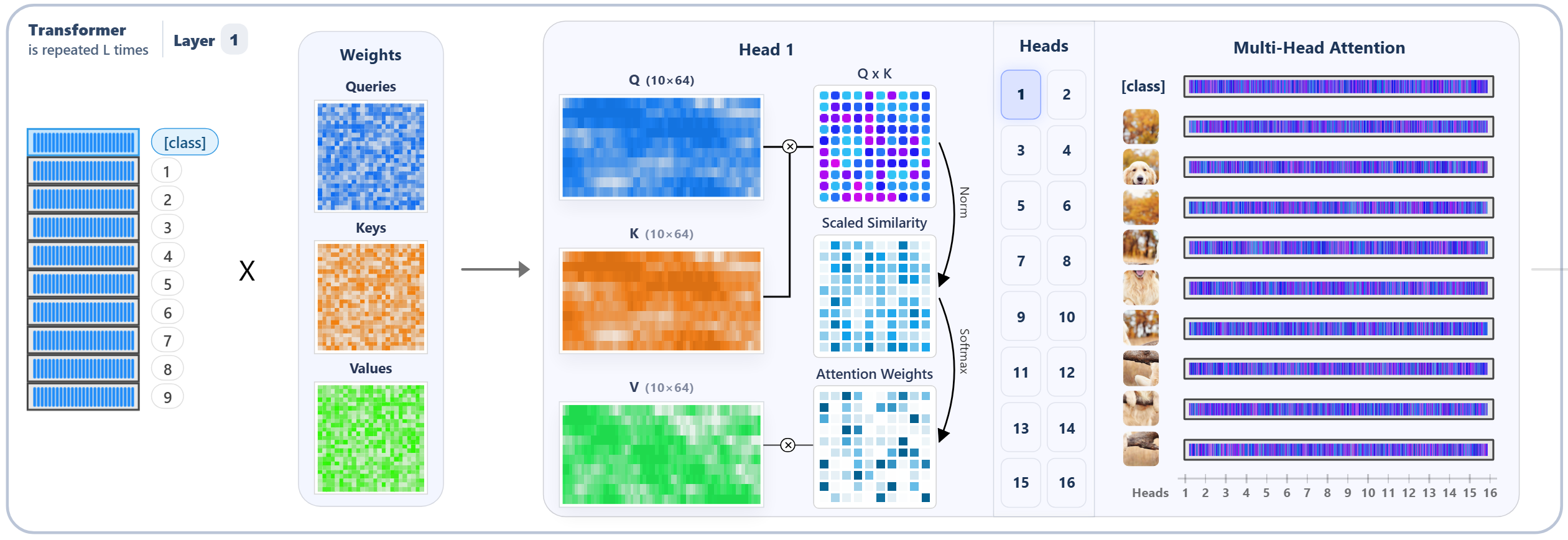}
    \caption{\textbf{Multi-Head Self-Attention:} ViT-Explainer animates the full attention computation to show how each head redistributes token information.}
    \label{fig:mha}
\end{figure}

Each encoder block is visualized as a sequence of animated transformations, including Layer Normalization, Multi-Head Self-Attention (Figure~\ref{fig:mha}), residual connections, and the MLP sublayer. Q, K, and V projections are rendered explicitly, followed by the dot-product interaction ($\mathbf{Q}\mathbf{K}^\top$), scaling, and softmax normalization. The transition from similarity scores to attention weights is animated row by row, highlighting how each token distributes attention across the sequence.

\begin{figure}[t]
    \centering
    \includegraphics[width=\columnwidth]{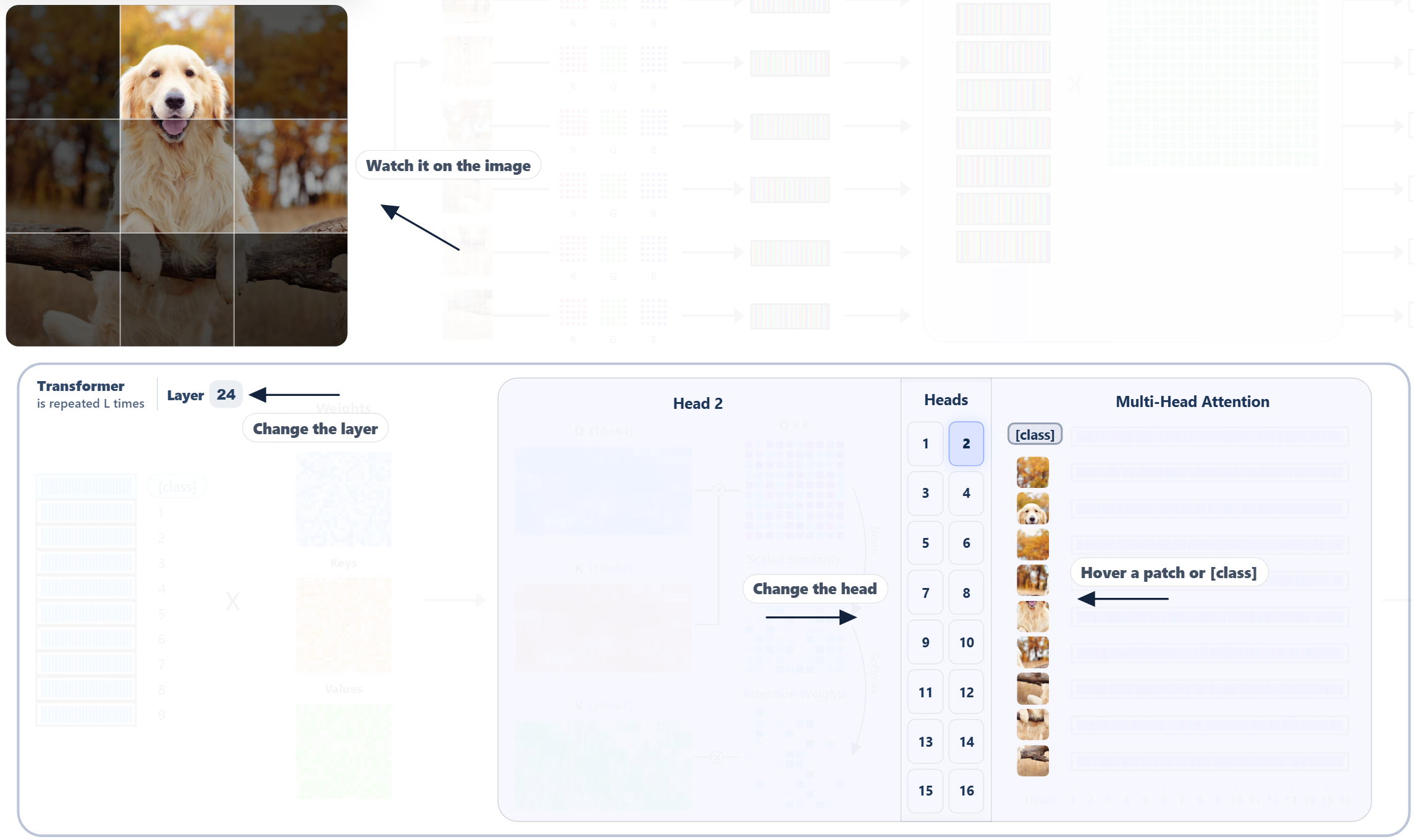}
    \caption{\textbf{Attention Mapping:} Visualizes how much attention a selected patch (or the [CLS] token) pays to other patches.}
    \label{fig:attention}
\end{figure}

Attention weights are displayed both as matrices and as overlays aligned with the image patches. Users can hover to inspect a token’s attention distribution (Figure~\ref{fig:attention}), switch between heads to compare learned patterns, and navigate layers to observe how spatial relationships evolve with depth. Residual additions and the two-stage MLP (W1, GELU, W2) are shown as explicit vector transformations.

\subsubsection{Classification and Logit Lens.}

\begin{figure}[t]
    \centering
    \includegraphics[width=\columnwidth]{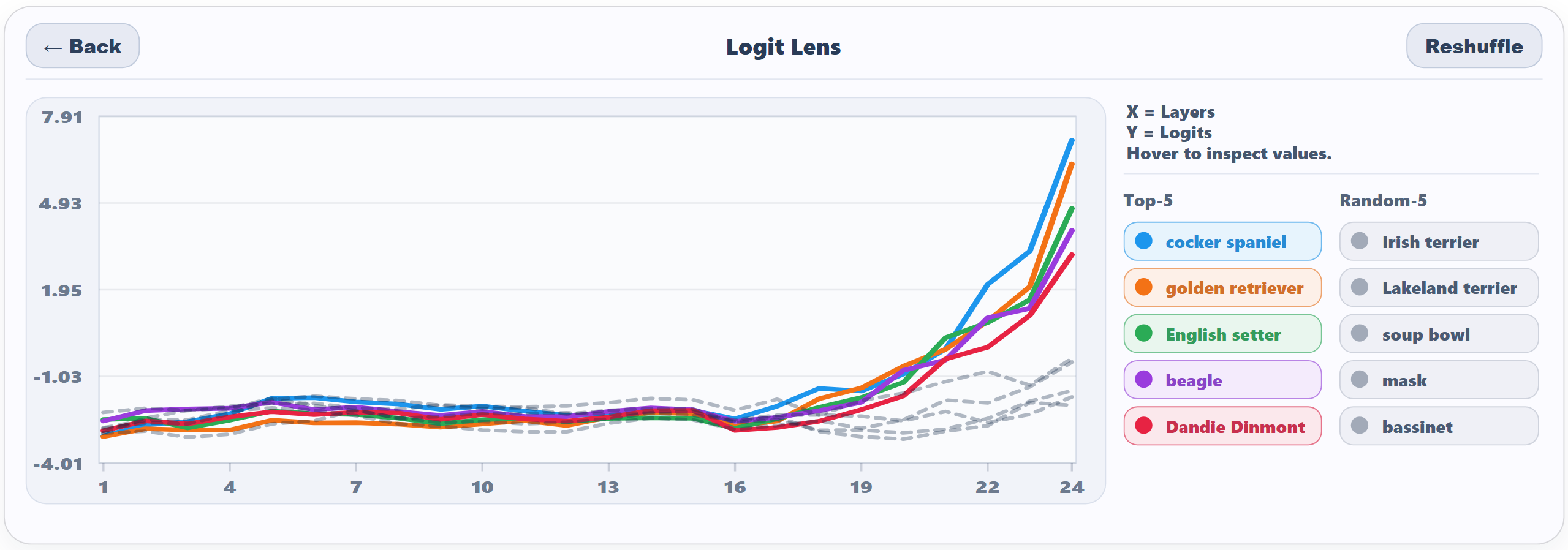}
    \caption{\textbf{Logit-Lens:} Layer-by-layer logit chart (colored curves for the top predicted classes) and a side panel to select and reshuffle classes for comparison.}
    \label{fig:logitlens}
\end{figure}

After the final layer, the CLS token is passed through the classification head to produce logits and softmax probabilities. A logit lens view (Figure~\ref{fig:logitlens}) applies the same head at every intermediate layer, visualizing how class scores evolve across depth and revealing the progressive sharpening of the model’s prediction \citep{belrose2023eliciting}.

\subsubsection{Interaction Modes.}
ViT-Explainer supports a guided walkthrough with synchronized explanations and animations, and a free exploration mode that allows direct interaction with heads, layers, and internal representations.

\section{User Evaluation}
\label{sec:eval}

To assess the usability and perceived effectiveness of ViT-Explainer, we conducted a user study with six participants.

\subsection{Participants and Procedure}

We recruited six participants (3 graduate and 3 undergraduate students from a Computer Science department) with varying familiarity with deep learning: \emph{beginner} ($N$=1; limited or no prior exposure to Transformers) and \emph{knowledgeable} ($N$=5; prior coursework or professional experience).

The study consisted of three phases: (1)~a pre-study questionnaire ($\sim$5 minutes) assessing self-reported familiarity with Transformer concepts; (2)~a guided exploration session ($\sim$20 minutes) in which participants walked through the complete ViT pipeline using a provided image, followed by free exploration with their own uploaded images; and (3)~a post-study questionnaire and semi-structured interview ($\sim$10 minutes), including usability ratings based on the System Usability Scale (SUS) \cite{brooke1996sus} and the NASA Task Load Index (NASA-TLX) \cite{hart1988development}, as well as open-ended feedback.

\subsection{Results}

\paragraph{Usability.} Participants reported low overall workload on the 1–7 NASA-TLX scale (M = 2.22) and high perceived performance (M = 6.33), suggesting that the system imposed limited cognitive burden during exploration. While Mental Demand was moderate (M = 4.00), other dimensions—Physical Demand (M = 1.33), Temporal Demand (M = 1.83), and Frustration (M = 1.17)—remained low, further indicating that interaction with the system was not cognitively taxing.

Consistent with these findings, perceived usability was high, with a mean SUS score of 90.42 (SD = 4.85). All participants scored above 85, reflecting a consistently positive and accessible user experience across varying levels of prior familiarity with Transformers.

\paragraph{Qualitative Feedback.} We conducted a thematic analysis \cite{clarke2017thematic} of the open-ended responses to identify perceived benefits and challenges of ViT-Explainer. Participants reported that ViT-Explainer was particularly helpful for understanding the attention mechanism at the patch level ($N$=5). Several noted that the interactive visualizations clarified aspects of Multi-Head Self-Attention that had remained abstract when learning from text-based resources alone. For example, P5 stated that the tool helped them \textit{“see how attention heads actually distribute focus across patches,”} making the computation more intuitive.

Participants appreciated the ability to inspect individual vectors and trace how representations propagate through successive stages of the architecture, connecting low-level embeddings to high-level predictions. As P3 noted, the system made abstract concepts \textit{“concrete and tangible,”} and the animated transitions helped them \textit{“build a mental model of the data flow.”}

Several participants noted that the step-by-step structure helped them understand how individual operations fit into the broader pipeline, rather than viewing patching, attention, and classification as isolated modules ($N$=4). As P2 explained, \textit{``Seeing each stage unfold step by step helped me understand how patching, attention, and classification connect to each other. Before, I thought of them as separate pieces, but the walkthrough made it clear how the output of one stage directly shapes the next.''}

Participants also highlighted the Logit Lens view as particularly insightful for understanding how predictions evolve across layers. Several noted that observing the gradual stabilization of class logits helped them see how intermediate representations progressively encode semantic information, rather than assuming that meaningful prediction emerges only at the final layer ($N$=3). As P1 explained, \textit{“Seeing the prediction evolve layer by layer made the depth make sense.”}

Despite the overall positive reception, participants identified usability issues. In particular, the interactive affordances were not always immediately apparent. Some users reported difficulty distinguishing interactive from non-interactive regions, leading to trial-and-error navigation. As P4 noted, \textit{“I wasn’t always sure what was clickable and what wasn’t, so I had to try a few things before figuring it out.”} This suggests a need for clearer visual cues (e.g., hover highlights or explicit UI indicators) to better guide user interaction.

\section{Limitations}
As discussed in §\ref{sec:architecture}, ViT-Explainer uses a pretrained FlexiViT-Large model adapted to a reduced $3 \times 3$ patch configuration for illustrative purposes. This trade-off reduces spatial granularity compared to standard ViT settings (e.g., $14\times14$ patches), and the model may correctly identify coarse semantic categories (e.g., ``dog'') but occasionally misclassify fine-grained distinctions (e.g., specific breeds). Attention dynamics at this resolution may also differ qualitatively from those in standard configurations, and future work should evaluate whether the patterns observed generalize to higher-resolution settings.

Our user evaluation involved six participants, predominantly with prior Transformer experience, from a single university. This small, relatively homogeneous sample limits the generalizability of the findings, particularly regarding the system's effectiveness for true beginners.

Finally, the system currently focuses on classification-oriented Vision Transformer models and does not yet extend to architectures designed for detection, segmentation, or multimodal tasks.

\section{Future Work}
\label{sec:future}

Future work focuses on improving usability, expanding model support, and strengthening evaluation. Based on user feedback, we are refining the intuitiveness of the free exploration mode by redesigning visual hints, such as hover indicators and highlights, to make interactive elements more discoverable while preserving visual clarity.

We also plan to extend ViT-Explainer to support additional Vision Transformer variants, requiring backend support for dynamic model selection and frontend adaptations for standard patch configurations (e.g., $14\times14$ or $16\times16$ grids). 

In parallel, we aim to conduct a new user study with a larger, more diverse pool of participants to better assess the system’s effectiveness across different levels of expertise. In the near future, we plan to release ViT-Explainer's source code and documentation to support broader community use and reproducibility.


\section{Conclusion}
\label{sec:conclusion}

We presented \textbf{ViT-Explainer}, a web-based interactive system that makes the internal computation of Vision Transformers transparent and explorable. The system provides an end-to-end visualization of the full ViT pipeline, integrating patch tokenization, Multi-Head Self-Attention, residual connections, MLP transformations, and layer-wise prediction tracing within a single coherent interface.

By combining a guided walkthrough with a free exploration mode, ViT-Explainer supports both structured learning and interactive inspection. Attention overlays and a vision-adapted logit lens allow users to observe how spatial relationships and class predictions evolve across layers.

Our user study indicates high usability and suggests that the system helps transform abstract Transformer operations into concrete, inspectable processes. We hope ViT-Explainer contributes to making Vision Transformer interpretability more accessible to students, educators, and researchers.

\section*{Ethics and Broader Impact}

The user study was reviewed and approved by the University of Notre Dame's Institutional Review Board (IRB), and all participants provided informed consent. We compensated participants with a USD 15 gift card.

ViT-Explainer is designed to increase transparency and accessibility in understanding Vision Transformer architectures. By exposing intermediate computations and prediction dynamics, the system aims to support responsible AI education and informed discussion about model behavior. We do not anticipate direct negative societal impacts from this work.

\section*{Acknowledgements}
This work was supported by the Pontificia Universidad Cat\'olica de Chile's School of Engineering under the 2025 Seed Award, the University of Notre Dame's College of Engineering Seed Award, and Notre Dame Global.

\bibliography{custom}

\appendix



\end{document}